%% file: main.tex
\documentclass[10pt,twocolumn,letterpaper]{article}

\usepackage{cvm}
\usepackage{times}
\usepackage{epsfig}
\usepackage{graphicx}
\usepackage{amsmath}
\usepackage{amssymb}

\usepackage{booktabs}
\usepackage{algorithm}
\usepackage{algorithmic}

\usepackage[pagebackref=true,breaklinks=true,letterpaper=true,colorlinks,bookmarks=false]{hyperref}

\newcommand{\keywords}[1]{{\bf \emph{Keywords: #1}}}

\cvmfinalcopy 

\ifcvmfinal\fi

\newcommand{\modelname}{SE-Mem\xspace}
\title{SE-Mem: Semantic-Episodic Memory for Scene-Consistent Generative View Synthesis}
\input{authors}
\begin{document}
\maketitle
\input{files/abstract}

\input{files/introduction}
\input{files/related}
\input{files/method}
\input{files/experiment}
\input{files/conclusion}
{\small
\bibliographystyle{cvm}
\bibliography{references}
}
\end{document}

%% file: authors.tex
\author{%
Changsong Lei\textsuperscript{1},
Yihan Li\textsuperscript{1},
Yu-Hui Wen\textsuperscript{2},
Huankun Sheng\textsuperscript{1}\\
Ming Li\textsuperscript{1},
Yeying Fan\textsuperscript{1},
Minnan Luo\textsuperscript{3},
Yong-Jin Liu\textsuperscript{1}\\[6pt]
{\small \textsuperscript{1}Tsinghua University}\\
{\small \textsuperscript{2}Beijing Jiaotong University}\\
{\small \textsuperscript{3}Xi'an Jiaotong University}\\[4pt]
{\tt\footnotesize \{leics23,liyihan23\}@mails.tsinghua.edu.cn}\\
{\tt\footnotesize \{shenghk,mingli\_thu,fanyeying,liuyongjin\}@tsinghua.edu.cn}\\
{\tt\footnotesize yhwen1@bjtu.edu.cn, minnluo@xjtu.edu.cn}%
}

%% file: files/abstract.tex
\begin{abstract}
Generative view synthesis aims to synthesize novel-view videos from reference observations along user-defined camera trajectories. Maintaining global scene consistency throughout this process remains challenging for camera-conditioned interactive video generators. Existing approaches typically select reference frames based solely on camera poses, overlooking whether the content remains semantically relevant or even visible. To address these limitations, we draw inspiration from semantic and episodic memory in human cognition, in which semantic memory captures generalized knowledge across experiences, whereas episodic memory preserves information tied to specific past events and contexts. Motivated by the complementary roles of semantic and episodic memory, we introduce SE-Mem, a Semantic-Episodic Memory framework that organizes historical observations in a structured 3D representation and augments its elements with semantic importance scores.
A two-stage memory retrieval strategy is designed to identify the most informative frames to provide contextual guidance. Taking into account both semantic significance and geometric visibility, SE-Mem preserves semantic coherence and fine-grained visual details during novel view synthesis. Extensive evaluations show that our approach improves visual quality and achieves competitive geometric consistency across multiple benchmarks. 
\end{abstract}

%% file: files/introduction.tex
\section{Introduction}

Novel View Synthesis (NVS) aims to generate visually realistic views from reference observations under novel camera viewpoints. While classical NVS focuses on reconstructing the observed scene, recent advances have extended this paradigm toward {\it generative view synthesis}~\cite{zhou2025stable}, where diffusion-based models synthesize novel-view videos along user-defined camera trajectories, enabling not only faithful reconstruction of observed content but also plausible expansion into unseen regions. 

\begin{figure*}[t]
    \centering
    \includegraphics[width=0.9\linewidth]{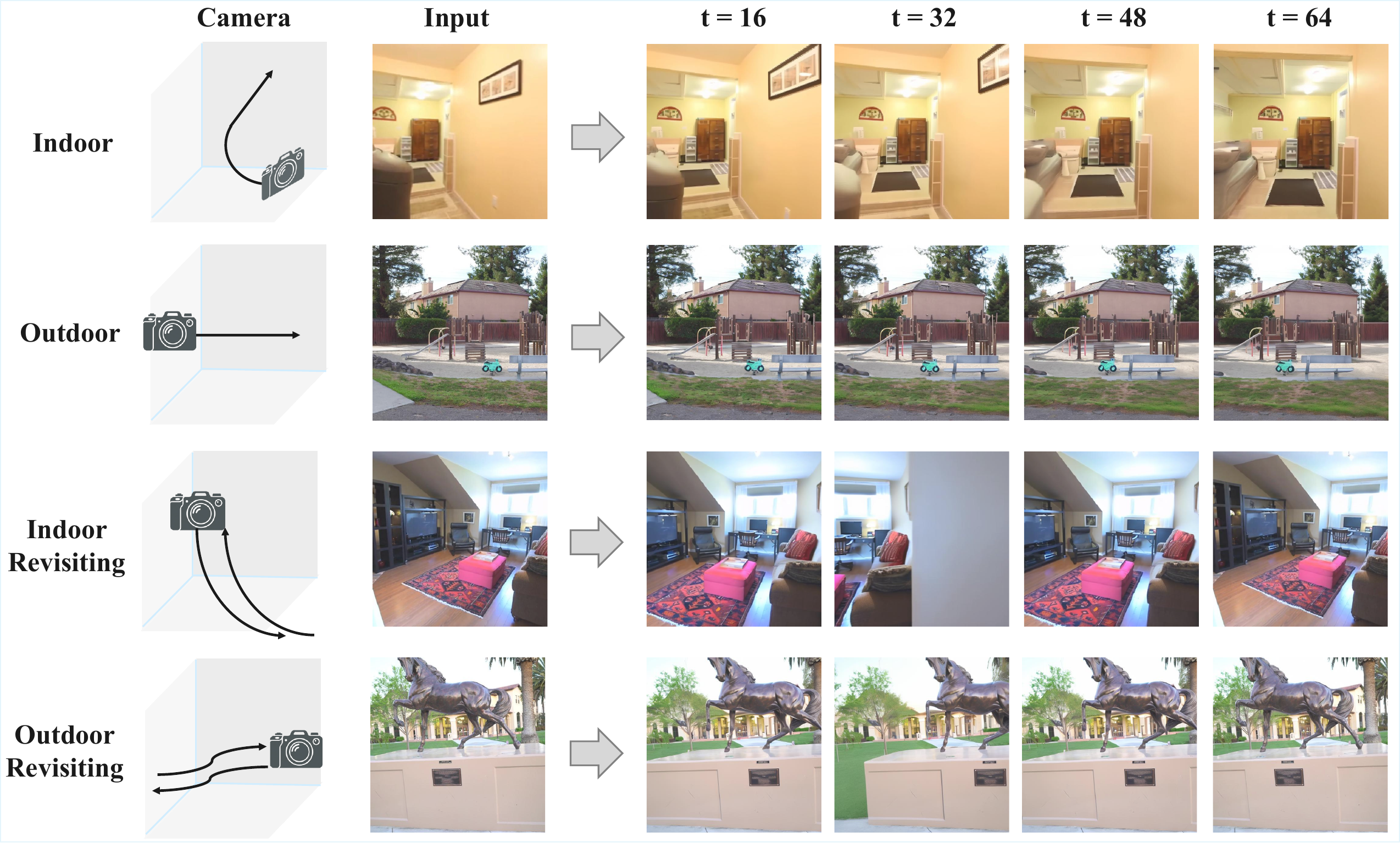}
    \caption{\textbf{Scene-consistent generative view synthesis with \modelname.}
    Given a single reference image and a user-defined camera trajectory, \modelname synthesizes a sequence of novel views while preserving previously observed scene content. Its Semantic-Episodic Memory combines structured 3D observations with semantic importance to retrieve relevant historical views, helping maintain scene consistency when the camera revisits earlier viewpoints in both indoor and outdoor environments.}
  \label{fig:teaser}
\end{figure*}

This generative setting introduces additional challenges in maintaining scene consistency along the camera trajectory, as illustrated in Fig.~\ref{fig:teaser}. Achieving visual fidelity and geometric consistency across complex camera trajectories requires a persistent scene representation that preserves previously observed content when viewpoints are revisited while incrementally incorporating new observations. Such consistency is important for interactive applications, including immersive gaming and embodied AI, where users or agents may revisit explored regions.

Recent efforts to enhance scene consistency can be broadly categorized into two paradigms. The first paradigm, {\it explicit 3D reconstruction-based methods} ~\cite{fridman2023scenescape,hu2026geometry,muller2024multidiff,ni2025recondreamer,ren2025gen3c,seo2024genwarp,yu2025wonderworld,yu2024viewcrafter}, synthesizes novel views through an iterative pipeline. These approaches typically construct 3D structures from input images, render visuals from new viewpoints, and refine the underlying 3D model with inpainted content. While effective within constrained regions, they are prone to drift due to accumulated errors in reconstruction, stitching, and inpainting. The second paradigm, {\it video-based methods}~\cite{ren2022look,wang2024motionctrl,xiao2025worldmem,yu2023long}, leverages generative priors to maintain scene consistency. These models, however, often struggle with limited context windows, making it difficult to preserve previously observed scene content under complex camera trajectories. Recent approaches mitigate this limitation by introducing 3D memory modules. Such modules enable the retrieval of spatially relevant keyframes from novel viewpoints, which reduces the memory and computational overhead associated with conditioning on a large number of frames~\cite{li2025vmem, yu2025context}. However, existing retrieval frameworks focus primarily on geometric co-visibility while overlooking semantic relevance. Consequently, large-scale background regions, which often provide limited information for conditional generation, are overrepresented among the selected keyframes merely due to their spatial extent.

In this paper, we propose SE-Mem, a Semantic-Episodic Memory framework designed to address the aforementioned limitations. We draw inspiration from cognitive memory theories that emphasize the interaction between semantic and episodic memory ~\cite{d2026genesis, tulving2002episodic, tulving1972episodic}. Specifically, the episodic memory is instantiated as a structured 3D point cloud that stores geometric cues as retrieval indices, while the semantic memory is encoded via semantic importance weights that reflect the informational value of each point. To build the episodic memory, we leverage an off-the-shelf point map estimator to incrementally construct the structured 3D point cloud throughout synthesis. To incorporate semantic awareness, we introduce a hierarchical importance scoring module that assigns semantic weights to individual points. 

Given a target camera pose, our framework identifies the most informative historical keyframes by accumulating semantic importance-weighted co-visibility scores, ensuring that the selected context is both spatially relevant and semantically meaningful.
However, relying solely on the most similar keyframes is often insufficient, as top-ranked candidates typically exhibit high mutual redundancy and may fail to capture all the information required for the target viewpoint. Such limitations often lead to incomplete scene coverage and a loss of fine-grained details in marginal regions. To address this, we introduce a complementary memory retrieval strategy that prioritizes informative regions absent from the primary context. Following the initial selection, our framework identifies supplementary frames that provide critical scene details required by the current pose but absent from the primary context frames. This two-stage strategy provides a more comprehensive context that helps address visibility gaps, resulting in spatially coherent and visually consistent synthesis.

The main contributions of this paper are summarized as follows:
\begin{itemize}
    \item We propose \modelname, a Semantic-Episodic Memory framework that retrieves historical keyframes containing semantically informative scene content to provide contextual guidance for scene-consistent generative view synthesis under a fixed conditioning-frame budget.
    \item Guided by the complementarity of semantic and episodic memory, we introduce a hierarchical importance scoring module that integrates semantic distinctiveness, object-level information density, and local structural cues, and associates the scores with a 3D episodic memory to prioritize informative scene elements over redundant background regions.
    \item We develop a two-stage memory retrieval strategy that incorporates a complementary mechanism to improve coverage of informative target-visible regions and select a more representative set of keyframes.
\end{itemize}

Experiments on the datasets of RealEstate10K, Tanks-and-Temples, and DL3DV evaluate visual fidelity and geometric consistency under general synthesis and viewpoint revisiting. Ablations examine the retrieval strategy, importance cues, and complementary-frame budget, while runtime and GPU-memory measurements characterize the computational cost.

%% file: files/related.tex
\section{Related Work}
\noindent\textbf{Novel View Synthesis} (NVS) aims to generate realistic and geometrically consistent views under novel camera viewpoints. Early approaches formulate NVS as a reconstruction and rendering problem, 
modeling scene geometry with representations such as NeRF~\cite{mildenhall2021nerf} and 3D Gaussian Splatting~(3DGS)~\cite{kerbl20233d}. While achieving high-fidelity rendering, these methods often suffer from sparse observations and large viewpoint variations due to incomplete scene coverage~\cite{gao2024gaussianflow,wang2025shape,wang2024dust3r}. Recent generative approaches alleviate these limitations by incorporating diffusion priors with geometric guidance, using intermediate representations such as point clouds~\cite{yu2024viewcrafter}, depth maps~\cite{ma2025you}, and 3D Gaussian primitives~\cite{yu2025wonderworld} to improve structural consistency.

Camera-controlled generation further enables view synthesis under user-defined trajectories. CameraCtrl~\cite{he2024cameractrl} introduces Plücker embeddings for camera conditioning, while MotionCtrl~\cite{wang2024motionctrl} leverages camera poses as global motion guidance. Other frameworks, including MultiDiff~\cite{muller2024multidiff}, ViewCrafter~\cite{yu2024viewcrafter}, and SEVA~\cite{zhou2025stable}, improve viewpoint controllability by exploiting geometric cues or sparse observations. However, these methods typically rely on limited contextual frames and lack explicit mechanisms for retaining relevant historical observations, which can lead to accumulated inconsistencies during extended camera trajectories.

\noindent\textbf{Memory-Augmented Video Generation.}
Video generators~\cite{blattmann2023stable,kong2024hunyuanvideo,wan2025wan,yang2024cogvideox,zhou2025stable} provide strong generative priors, but fixed-window synthesis offers limited access to historical observations. Structured memory and retrieval mechanisms address this limitation by retaining scene information beyond the active conditioning window. Existing approaches mainly rely on either geometry-guided generation or memory retrieval. Geometry-guided methods~\cite{hu2026geometry,ren2025gen3c,wu2025video,yu2025wonderworld,yu2024viewcrafter} exploit explicit 3D representations, such as point clouds and Gaussian primitives, to provide structural guidance, but their performance depends on the quality of reconstructed geometry. To mitigate errors from global reconstruction, AnchorWeave~\cite{wang2026anchorweave} retrieves local geometric memories according to target-trajectory coverage and reconciles them through a multi-anchor controller. Retrieval-based methods~\cite{li2025vmem,sun2025worldplay,yu2025context} select past observations using spatial structures or visibility cues, enabling memory augmentation without modifying the generator. Memorize When Needed~\cite{guo2026memorize} further decouples memory control from generation, selectively incorporating spatially relevant historical information through camera-aware gating. However, geometric retrieval may fail to identify observations containing semantically important regions.

%% file: files/method.tex
\section{Method}
\label{sec:method}

Let $x_0 \in \mathbb{R}^{H \times W \times 3}$ denote the initial reference image with its corresponding camera parameters $c_0$. Given a user-defined camera trajectory $\mathcal{C}=\{c_t\}_{t=1}^{T}$, our goal is to synthesize the corresponding video sequence $\mathcal{V}=\{x_t\}_{t=1}^{T}$. We build upon a camera-conditioned video generator that follows an autoregressive synthesis process, where each generation step produces a segment of $M$ frames conditioned on historical observations and their corresponding camera parameters.

\begin{figure*}[t]
    \centering
    \includegraphics[width=0.9\linewidth]{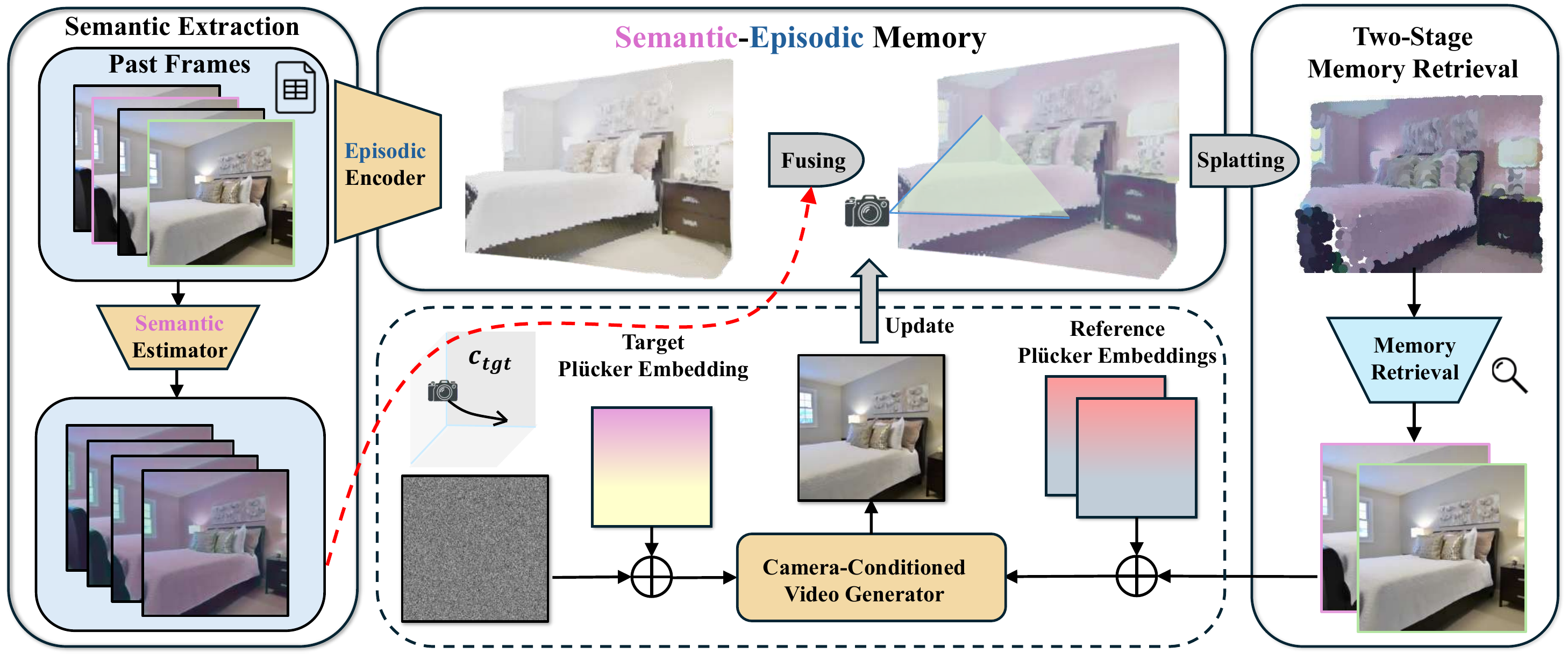}
    \caption{\textbf{Overview of \modelname.} 
    Given a sequence of target camera poses $\{c_{tgt}\}$, \modelname maintains a Semantic-Episodic Memory that combines 3D episodic observations with semantic importance scores. Historical frames are processed to obtain 3D point representations and importance maps, which are integrated into a voxelized memory representation. During synthesis, a two-stage retrieval strategy selects reference frames by considering target-view visibility and voxel importance. The retrieved frames and their corresponding camera conditions are then provided to a camera-conditioned video generator to synthesize novel views, while newly generated observations are incrementally incorporated into the memory.
    }
    
    \label{fig:framework}
\end{figure*}

To enable consistent synthesis beyond a limited temporal context, we propose \modelname, a Semantic-Episodic Memory framework illustrated in Fig.~\ref{fig:framework}. Rather than relying on a fixed set of recent frames, \modelname constructs a structured 3D episodic memory from previously observed scene content and assigns semantic importance scores to its 3D elements. By combining episodic geometry with semantic relevance, it retrieves historical views that provide both geometrically relevant and visually informative context for subsequent synthesis. Our two-stage retrieval first selects historical frames relevant to the target viewpoint and then supplements them with observations containing missing scene details. The following sections describe semantic importance scoring, episodic memory construction, and the two-stage retrieval mechanism. The conditioning budget is fixed at $K$ historical frames, so retrieval must balance target-view relevance with coverage of informative scene regions.

\subsection{Semantic Importance Scoring}
\label{sec:semantic}

\begin{figure*}[t]
    \centering
    \includegraphics[width=1.0\linewidth]{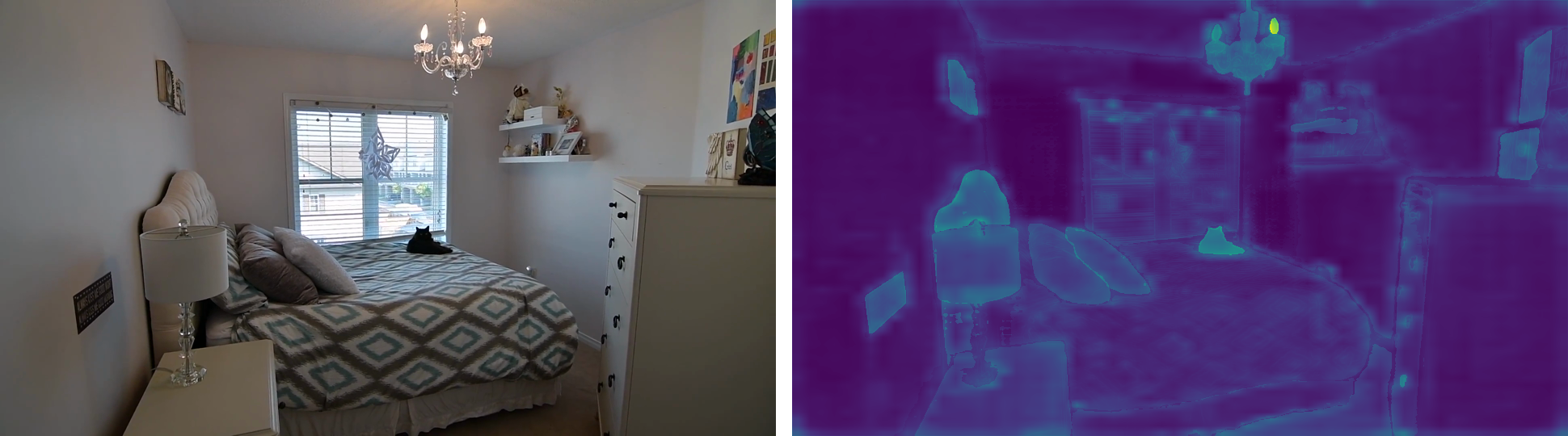}
    \caption{Visualization of importance maps. Given an input scene (left), our hierarchical importance scoring module (Eq.~\eqref{Eq:importance}) produces an importance map (right),  where brighter regions indicate higher importance values. This map guides our \modelname to prioritize semantically informative scene elements during both memory construction and retrieval.}
    
    \label{fig:importance}
\end{figure*}

Human perception naturally prioritizes semantically informative visual cues rather than treating all observations equally~\cite{tulving1972episodic}. Inspired by this principle, we assign importance scores to observed regions using three complementary cues: semantic distinctiveness, object-level information density, and local structural details. As illustrated in Fig.~\ref{fig:importance}, these cues capture visually informative regions that may be overlooked by geometry-based retrieval and are associated with corresponding 3D points to guide memory construction and retrieval.

\noindent\textbf{Semantic Distinctiveness Extraction.}
Given a synthesized frame $x_t \in \mathbb{R}^{H \times W \times 3}$, we extract dense visual features using the pretrained SAM2 vision encoder~\cite{ravi2025sam}. Prior studies have shown that pretrained visual representations encode spatially localized semantic information, while distances in these feature spaces can reflect perceptual and semantic differences between visual regions~\cite{caron2021emerging,oquab2024dinov2,zhang2023tale}. Based on this property, we measure the distinctiveness of each region by its deviation from the dominant scene feature distribution:
\begin{equation}
S_{sem}(i)=||F_i-\bar{F}||_2,
\end{equation}
where $F_i$ denotes the feature vector at spatial location $i$ and $\bar{F}$ represents the mean feature embedding over the frame. Regions with larger deviations indicate distinctive visual patterns and receive higher distinctiveness scores. The resulting map is upsampled to the original resolution for subsequent fusion.

\noindent\textbf{Object-Level Information Density Scoring.}
While semantic distinctiveness measures the difference between a region and the overall scene appearance, it does not capture the amount of detailed information contained within an object. We therefore estimate object-level information density based on internal visual variations. Specifically, object masks are used to define object regions, and the information density of each region is computed as:
\begin{equation}
S_{obj}(i)=\frac{\sigma(F_i)}{A_i},
\end{equation}
Here, $i$ indexes an object region, $A_i$ denotes its area, and $\sigma(F_i)$ represents the average feature variance within the region. The region score is assigned to its corresponding image pixels for fusion with the other cues. Objects with larger internal feature variations typically contain richer visual details, while area normalization measures the amount of variation per unit area. Regions with higher information density are assigned larger importance scores.

\noindent\textbf{Local Structural Extraction.}
To preserve fine-grained visual details that may not be captured by semantic features, we introduce a local structural cue based on image gradients. Specifically, the Sobel operator is used to compute gradient magnitude:
\begin{equation}
G=\sqrt{G_x^2+G_y^2},
\end{equation}
which is normalized to obtain the local structural importance map $S_{loc}$.

Finally, the three cues are fused into an importance map $I$:
\begin{equation}
\label{Eq:importance}
I=\lambda_1S_{sem}+\lambda_2S_{obj}+\lambda_3S_{loc},
\end{equation}
where $\lambda_1$, $\lambda_2$, and $\lambda_3$ control the contribution of each component. The resulting importance values are associated with corresponding 3D points during memory construction and update for importance-aware retrieval.

\subsection{Semantic-Episodic Memory}
\label{sec:ssmem}

At generation step $s+1$, given previously synthesized frames $\{x_t\}_{t=0}^{sM}$ and their corresponding camera trajectories $\{c_t\}_{t=0}^{sM}$, our goal is to retrieve historical observations that provide informative context for the target poses $\{c_t\}_{t=sM+1}^{(s+1)M}$.  Guided by the cognitive principle that semantic knowledge encodes high-level informational salience while episodic memory preserves  spatial context, we design a memory representation that preserves previously observed scene content while prioritizing informative visual elements for future retrieval.

Specifically, we process each historical frame with CUT3R~\cite{wang2025continuous} to obtain a point map $P_t \in \mathbb{R}^{H \times W \times 3}$ and a corresponding confidence map $C_t \in \mathbb{R}^{H \times W \times 1}$. The point map provides the geometric representation of the observed scene, while the confidence map measures the reliability of the reconstructed structure. Meanwhile, we obtain the importance map $I_t$ using Eq.~\eqref{Eq:importance}. Each pixel observation is therefore associated with geometric, semantic, and temporal attributes:
\begin{equation}
\left(P_t^{(u,v)}, C_t^{(u,v)}, I_t^{(u,v)}, t\right),
\end{equation}
where $P_t^{(u,v)}$ denotes the 3D location, $C_t^{(u,v)}$ represents geometric reliability, $I_t^{(u,v)}$ indicates the importance of the observed content, and $t$ records the source observation.

However, directly accumulating 3D points and their attributes from all synthesized frames leads to severe memory redundancy, as the same spatial regions are repeatedly observed from different viewpoints. To maintain a compact yet persistent scene memory, we aggregate accumulated 3D points and their attributes into a voxel-based representation. 

Specifically, we discretize the continuous 3D space into a fixed-resolution voxel grid with a predefined voxel size $b$. Each reconstructed point is assigned to its corresponding voxel according to its 3D coordinate. With fixed voxelization, the number of geometric records is determined by occupied voxels rather than accumulated observations. For each voxel $V_k$, we aggregate observations falling into the same spatial region and maintain the most reliable geometric observation, the associated importance score, and all historical frames that have contributed observations to this voxel.

Formally, after aggregating all 3D points and their attributes, the Semantic-Episodic Memory is constructed as a set of voxels:
\begin{equation}
\mathcal{M}
=
\{V_k=(X_k,C_k,I_k,\mathcal{F}_k)\mid k\in\mathcal K\},
\end{equation}
where $\mathcal K$ denotes the set of occupied voxel indices. $X_k$ represents the spatial location of voxel $k$, $C_k$ denotes the confidence of the most reliable observation retained in the voxel, $I_k$ represents the semantic importance associated with this voxel, and $\mathcal F_k$ stores the historical frames that have observed this spatial region.

The voxel-based representation merges repeated observations of the same region instead of storing their geometry independently. During generation, newly synthesized frames are incrementally integrated by updating voxel attributes and expanding their associated observation history. The number of geometric records is therefore determined by occupied voxels rather than accumulated point observations. Exploring new regions can introduce additional voxels, and the frame-association sets can grow as observations accumulate; voxelization does not by itself imply a constant bound on total memory. The fixed context budget $K$ instead controls the number of historical frames passed to the generator.

\noindent\textbf{Geometry as a Retrieval Index.}
The memory separates spatial indexing from visual conditioning. Voxel locations, confidence scores, and importance values determine which historical observations are retrieved, while the generator receives the selected RGB frames and their camera parameters. We do not render the accumulated point cloud into conditioning images or replace stored frame content with reconstructed appearance. Inaccurate geometry therefore does not directly alter the pixels of a retained observation. Historical image evidence remains available independently of its 3D reconstruction, while geometry supplies the spatial associations for querying it from a new viewpoint. Section~\ref{sec:discussion} discusses the consequences of inaccurate indexing.

\subsection{Memory Retrieval}
\label{sec:retrieval}

Given a target camera pose $c_{\mathrm{tgt}}$, we retrieve a compact set of historical frames from the Semantic-Episodic Memory. Each voxel stores its spatial location, semantic importance, and source-frame associations, allowing frames to be scored by their support for the target view. Since a single ranking may favor redundant frames covering similar regions, we adopt a two-stage strategy: primary retrieval ranks frames by their overall target-view support, while complementary retrieval adds frames covering informative regions not represented in the primary context.

\subsubsection{Primary Retrieval}
\label{sec:primary}
To select relevant historical frames, we splat the 3D memory $\mathcal{M}$ into the target camera space. Voxels are processed in front-to-back order according to their target-view depth, allowing spatial occlusions to be modeled through a pixel-wise transmittance map $\mathcal{T}$. We initialize $\mathcal{T}(u,v)=1$ at each target-image pixel. For a voxel $V_k$ projected to $(u,v)$, its confidence $C_k$ is normalized to obtain an opacity $\alpha_k$. Here, $C_k$ is the CUT3R confidence of the most reliable observation retained in the voxel, consistent with the memory definition in Section~\ref{sec:ssmem}.

The contribution of voxel $V_k$ is
\begin{equation}
\label{eq:voxel_vote}
w_k=\mathcal{T}_k\alpha_k I_k,
\end{equation}
where $\mathcal{T}_k=\mathcal{T}(u,v)$ is evaluated before incorporating this voxel. We then accumulate its contribution over the historical frames associated with it:
\begin{equation}
\label{eq:primary_score}
S_t=\sum_{k:t\in\mathcal{F}_k}w_k.
\end{equation}
The transmittance map is updated within the local splatting neighborhood $\mathcal{N}_r(u,v)$:
\begin{equation}
\label{eq:transmittance}
\mathcal{T}(x,y)\leftarrow\mathcal{T}(x,y)(1-\alpha_k),
\quad (x,y)\in\mathcal{N}_r(u,v).
\end{equation}
Here, $\mathcal{N}_r(u,v)$ denotes the pixel neighborhood of radius $r$ around the projected voxel center. The update accounts for the footprint covered by each splat. Transmittance summarizes attenuation by previously processed voxels, reducing the contribution of observations behind already covered regions. Voxels whose projected transmittance falls below the threshold $\epsilon$ are skipped, as detailed in Algorithm~\ref{alg:retrieval}.

The resulting frame score measures importance-weighted geometric support for the target view. Confidence weighting reduces the contribution of unreliable reconstructed observations, while the front-to-back transmittance update accounts for viewpoint-dependent occlusion. Frames receive higher scores when they observe voxels that are both reliable and informative for the target view.

\begin{algorithm}[!t]
\caption{Primary Retrieval Algorithm}
\label{alg:retrieval}
\begin{algorithmic}[1]
\renewcommand{\algorithmicrequire}{\textbf{Input:}}
\renewcommand{\algorithmicensure}{\textbf{Output:}}
\REQUIRE 3D Memory $\mathcal{M} = \{V_k = (X_k,
C_{k}, I_{k}, \mathcal{F}_{k})\mid k \in \mathcal{K} \}$, Target Camera Parameters $c_{\mathrm{tgt}}$
\ENSURE Frame scores $\{S_t\}$
\STATE Initialize $S_t \leftarrow 0$ for all historical frames $t$
\STATE Initialize transmittance map $\mathcal{T}(u, v) \leftarrow 1.0$ for all pixels $(u, v)$
\STATE Sort voxels $\{V_k\}$ in \textit{front-to-back} order based on their depth $d_k$
\FOR{each voxel $V_k$ in sorted set}
    \STATE Project $V_k$ to pixel $(u, v)$ given $c_{\mathrm{tgt}}$
    \IF{$\mathcal{T}(u, v) < \epsilon$}
        \STATE \textbf{continue}
    \ENDIF
    \STATE $\alpha_k \leftarrow \text{Normalize}(C_k)$
    \STATE $w_k \leftarrow \mathcal{T}(u, v) \cdot \alpha_k \cdot I_k$
    \FOR{each frame index $t \in \mathcal{F}_k$}
        \STATE $S_t \leftarrow S_t + w_k$
    \ENDFOR
    \STATE $\mathcal{T}(x, y) \leftarrow \mathcal{T}(x, y) \cdot (1 - \alpha_k), \forall (x, y) \in \mathcal{N}_r(u, v)$
\ENDFOR
\RETURN $\{S_t\}$
\end{algorithmic}
\end{algorithm}

\subsubsection{Complementary Retrieval}
\label{sec:complementary}
Primary retrieval ranks historical frames by their aggregate support, but highly ranked frames may observe overlapping regions. Under a context budget of $K$ frames, we reserve $N$ positions for complementary retrieval and select the top $K-N$ primary frames as $\mathcal{I}_p$. This allocation preserves the total budget while allowing additional observations to address gaps in the primary context.

Let $\mathcal{K}_{\mathrm{vis}}$ denote the indices of voxels that project within the target-image region and are visible under the target viewpoint. The uncovered voxel-index set is
\begin{equation}
\label{eq:uncovered}
\mathcal{U}=\{k\in\mathcal{K}_{\mathrm{vis}}\mid
\mathcal{F}_k\cap\mathcal{I}_p=\emptyset\}.
\end{equation}
For each remaining candidate frame $t\notin\mathcal{I}_p$, the complementary score is
\begin{equation}
\label{eq:complementary_score}
S'_t=\sum_{k\in\mathcal{U},\,t\in\mathcal{F}_k} I_k.
\end{equation}
Algorithm~\ref{alg:complementary} computes these scores from the target-visible voxels and the primary frame set. Unlike ranking by visible area alone, this score favors observations of high-importance regions absent from the primary context.

\begin{algorithm}[!t]
\caption{Complementary Retrieval Algorithm}
\label{alg:complementary}
\begin{algorithmic}[1]
\renewcommand{\algorithmicrequire}{\textbf{Input:}}
\renewcommand{\algorithmicensure}{\textbf{Output:}}
\REQUIRE Visible voxel set $\{V_k\mid k\in\mathcal{K}_{\mathrm{vis}}\}$ under target view $c_{\mathrm{tgt}}$, Primary indices $\mathcal{I}_p$
\ENSURE Frame scores $\{S_{t}'\}$
\STATE Initialize uncovered voxel-index set $\mathcal{U} \leftarrow \emptyset$ 
\FOR{each visible voxel index $k\in\mathcal{K}_{\mathrm{vis}}$}
    \IF{$\mathcal{F}_k \cap \mathcal{I}_p = \emptyset$}
        \STATE $\mathcal{U} \leftarrow \mathcal{U} \cup \{k\}$
    \ENDIF
\ENDFOR
\STATE Initialize $S_t' \leftarrow 0$ for all $t$
\FOR{each uncovered voxel index $k\in\mathcal{U}$}
    \FOR{each frame index $t \in (\mathcal{F}_k \setminus \mathcal{I}_p)$}
        \STATE $S_t' \leftarrow S_t' + I_k$
    \ENDFOR
\ENDFOR
\RETURN $\{S_t'\}$
\end{algorithmic}
\end{algorithm}

We select the top $N$ remaining frames according to $S'_t$ to form $\mathcal{I}_c$. The final context $\mathcal{I}_p\cup\mathcal{I}_c$ contains $K$ historical frames, together with their camera parameters, for conditioning the next synthesis step. The complementary stage improves representation of informative target-visible regions without increasing the conditioning-frame budget. Its effect and the tradeoff between primary and complementary context are evaluated in Section~\ref{sec:ablation_n}.

%% file: files/experiment.tex
\section{Experiments}
\label{sec:experiments}

\subsection{Experimental Setup}

\textbf{Implementation Details.} 
Following VMem~\cite{li2025vmem}, we set the context size to $K=4$ and the segment length to $M=4$. The importance fusion weights are set to $\lambda_1=0.4$, $\lambda_2=0.5$, and $\lambda_3=0.1$ for semantic distinctiveness, object-level information density, and local structure, respectively. To ensure compatibility with the SEVA backbone, all frames are resized and center-cropped to $576\times576$. This resolution is specific to the SEVA configuration. The retrieval module selects frame indices and leaves image preprocessing to the chosen backbone, so it does not impose a common output resolution across generators. We fine-tune the model with LoRA on eight A100 GPUs for 30,000 iterations. During inference, we select $K-N=3$ primary frames and $N=1$ complementary frame. The complementary-frame budget is analyzed in Section~\ref{sec:ablation_n}, and memory usage and runtime are evaluated in Section~\ref{sec:efficiency}.

\noindent\textbf{Datasets.}
We evaluate our method on two widely used datasets for novel view synthesis: RealEstate10K~\cite{zhou2018stereo}, which contains monocular videos with diverse camera trajectories, and Tanks-and-Temples~\cite{knapitsch2017tanks}, which includes large-scale indoor and outdoor scenes with substantial viewpoint variations. We additionally evaluate generalization on DL3DV~\cite{ling2024dl3dv} and on in-the-wild images collected from internet sources (Section~\ref{sec:generalization}). For DL3DV, we use the 11 sample scenes provided by the dataset, randomly select 10 starting frames per scene, and construct one evaluation clip per starting position, yielding 110 clips.

\noindent\textbf{Evaluation.}
Following the evaluation protocol of VMem~\cite{li2025vmem}, we assess the generated views from both visual and geometric perspectives. We use PSNR and SSIM~\cite{wang2004image} to measure pixel-level fidelity and structural similarity, respectively, and LPIPS~\cite{zhang2018unreasonable} to evaluate perceptual similarity. For geometric evaluation, we estimate camera poses from the generated frames and compute the rotation distance $R_{\mathrm{dist}}$ and translation distance $T_{\mathrm{dist}}$ with respect to the target camera trajectory.

The metrics provide complementary evidence. SSIM measures structural agreement at corresponding image locations and can decrease with spatial misalignment~\cite{wang2004image}. LPIPS compares perceptual image features~\cite{zhang2018unreasonable}, while pose errors assess geometric agreement with the prescribed trajectory. Perceptual improvement alone therefore does not establish accurate camera control. Evaluating the return trajectory additionally tests recovery of previously observed content, beyond generating plausible views of newly explored regions.

\subsection{Synthesis and Scene Consistency}
\label{sec:results}
Starting from a single reference image, we evaluate \modelname under diverse camera trajectories, including general synthesis and viewpoint revisiting scenarios. These settings assess short-trajectory visual fidelity and the preservation of previously observed content during longer camera exploration, respectively.

\subsubsection{General Synthesis}
For the general synthesis task, we generate a 32-frame sequence along a predefined camera trajectory with a temporal stride of 1. Metrics are computed over all generated frames and reported in Table~\ref{tab:general}. \modelname achieves competitive performance on both datasets, demonstrating strong visual fidelity under different camera movements. In particular, its consistent LPIPS improvement suggests that semantic-aware memory provides informative historical context for preserving perceptually relevant scene details. The geometric metrics show a more mixed comparison: \modelname yields the lowest rotation error on RealEstate10K, while some baselines achieve lower translation error or better geometric metrics on Tanks-and-Temples.

\begin{table*}[t]
\centering
\caption{Quantitative comparison for the General Synthesis task on RealEstate10K and Tanks-and-Temples datasets. Best results are highlighted in bold, and second-best results are underlined.}
\label{tab:general}
\resizebox{\linewidth}{!}{
\begin{tabular}{l ccccc ccccc}
\toprule
 & \multicolumn{5}{c}{\textbf{RealEstate10K}} & \multicolumn{5}{c}{\textbf{Tanks-and-Temples}} \\
\cmidrule(lr){2-6} \cmidrule(lr){7-11}
Method & PSNR $\uparrow$ & SSIM $\uparrow$ & LPIPS $\downarrow$ & $R_{dist} \downarrow$ & $T_{dist} \downarrow$
& PSNR $\uparrow$ & SSIM $\uparrow$ & LPIPS $\downarrow$ & $R_{dist} \downarrow$ & $T_{dist} \downarrow$ \\
\midrule
MotionCtrl~\cite{wang2024motionctrl}
& 12.54 & 0.353 & 0.565 & 0.813 & 0.742
& 11.76 & 0.430 & 0.656 & 1.280 & 0.796 \\

ViewCrafter~\cite{yu2024viewcrafter}
& 16.57 & 0.583 & 0.447 & 0.474 & 0.623
& 12.84 & 0.461 & 0.598 & 1.039 & 0.820 \\

GEN3C~\cite{ren2025gen3c}
& \underline{18.21} & \textbf{0.664} & 0.372 & 0.505 & \underline{0.368}
& 14.27 & 0.490 & 0.541 & \textbf{0.554} & 0.701 \\

SEVA~\cite{zhou2025stable}
& 18.09 & 0.641 & \underline{0.349} & \underline{0.467} & \textbf{0.293}
& 14.50 & \underline{0.522} & 0.539 & \underline{0.601} & 0.605 \\

VMem~\cite{li2025vmem}
& 17.98 & 0.632 & 0.356 & 0.592 & 0.481
& \underline{14.73} & 0.514 & \underline{0.513} & 0.715 & \textbf{0.472} \\

\midrule
\textbf{\modelname (Ours)}
& \textbf{18.46} & \underline{0.647} & \textbf{0.335} & \textbf{0.407} & 0.432
& \textbf{15.25} & \textbf{0.541} & \textbf{0.489} & 0.633 & \underline{0.548} \\

\bottomrule
\end{tabular}
}
\end{table*}

\subsubsection{Viewpoint Revisiting}

\begin{figure*}[t]
    \centering
    \includegraphics[width=1.0\linewidth]{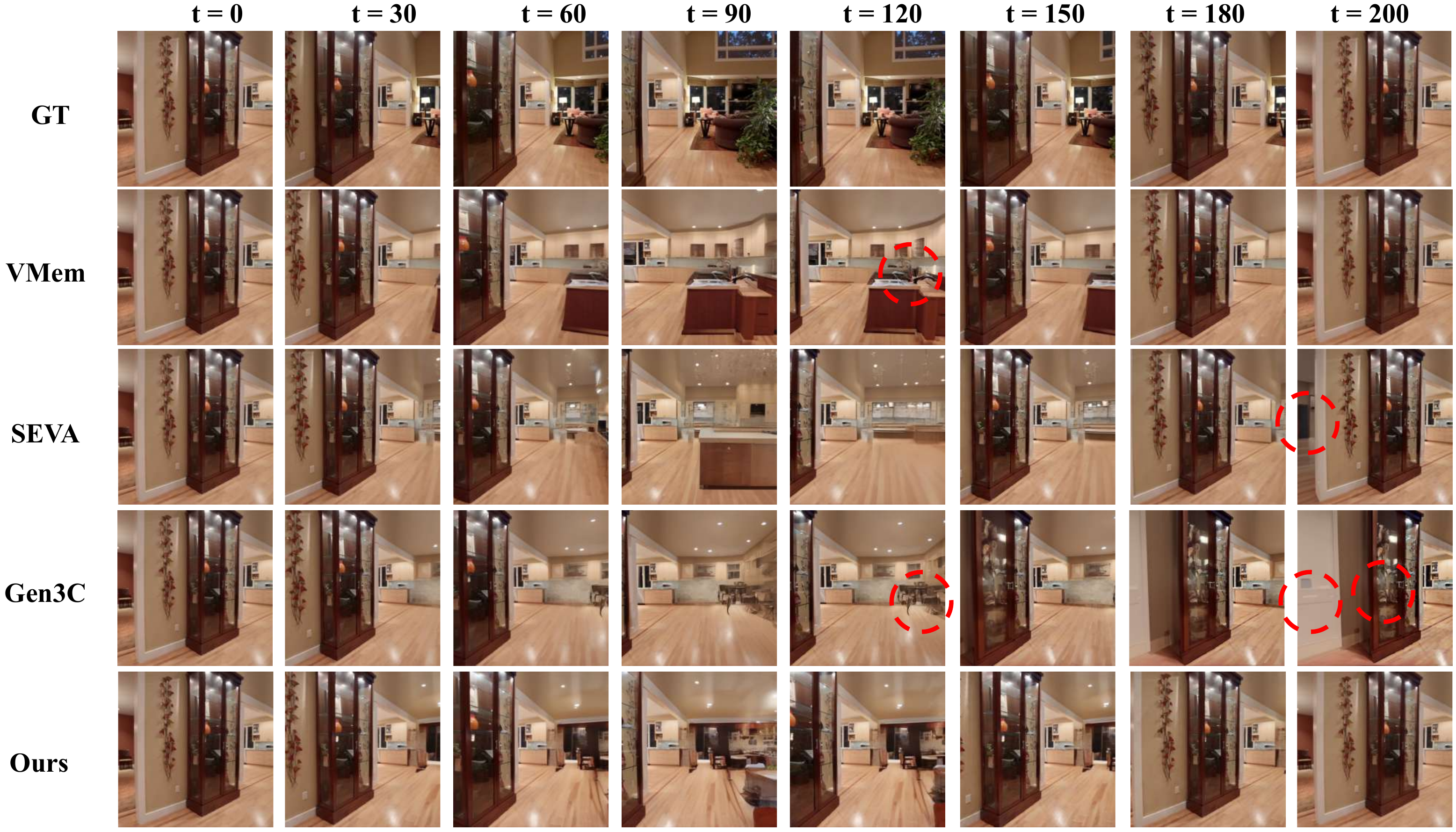}
    \caption{Qualitative results on the Viewpoint Revisiting task. Our method maintains scene consistency when returning to previously observed viewpoints, successfully recovering the original environment with reduced perceptual drift and improved visual fidelity.}
    
    \label{fig:comparison}
\end{figure*}
To evaluate the ability of \modelname to preserve scene consistency under viewpoint revisiting, we construct a forward-and-back trajectory. The camera first moves forward for 100 frames and then retraces the same path back to the initial viewpoint, forming a 200-frame sequence sampled at a temporal stride of 5. This setting requires the model to recover previously observed scene content after traversing intermediate viewpoints, providing a direct evaluation of memory-based scene preservation. We compute metrics on all frames generated during the return trajectory, and the results are summarized in Table~\ref{tab:revisit}.

\begin{table*}[t]
\centering
\caption{Quantitative results on the Viewpoint Revisiting task on RealEstate10K and Tanks-and-Temples datasets. Best results are highlighted in bold, and second-best results are underlined.}
\label{tab:revisit}
\resizebox{\linewidth}{!}{
\begin{tabular}{l ccccc ccccc}
\toprule
 & \multicolumn{5}{c}{\textbf{RealEstate10K}} & \multicolumn{5}{c}{\textbf{Tanks-and-Temples}} \\ 
\cmidrule(lr){2-6} \cmidrule(lr){7-11} 
Method & PSNR $\uparrow$ & SSIM $\uparrow$ & LPIPS $\downarrow$ & $R_{dist} \downarrow$ & $T_{dist} \downarrow$ & PSNR $\uparrow$ & SSIM $\uparrow$ & LPIPS $\downarrow$ & $R_{dist} \downarrow$ & $T_{dist} \downarrow$ \\
\midrule
MotionCtrl~\cite{wang2024motionctrl}
& 9.78 & 0.314 & 0.660 & 1.407 & 0.920
& 8.15 & 0.265 & 0.792 & 2.854 & 1.826 \\

SEVA~\cite{zhou2025stable}
& 13.20 & 0.501 & 0.459 & 0.967 & 0.516
& 11.09 & 0.329 & 0.636 & 1.973 & 0.921 \\

GEN3C~\cite{ren2025gen3c}
& 13.09 & 0.486 & 0.472 & \underline{0.871} & \textbf{0.461}
& 11.51 & 0.325 & 0.649 & 1.564 & 1.387 \\

ViewCrafter~\cite{yu2024viewcrafter}
& 13.38 & 0.473 & 0.434 & 0.912 & 0.654
& 11.35 & 0.309 & 0.672 & 1.746 & \textbf{0.710} \\

VMem~\cite{li2025vmem}
& \underline{13.64} & \underline{0.525} & \underline{0.418} & 1.032 & 0.547
& \underline{12.10} & \textbf{0.342} & \underline{0.627} & \underline{1.529} & 0.852 \\

\midrule
\textbf{\modelname (Ours)}
& \textbf{14.22} & \textbf{0.549} & \textbf{0.387} & \textbf{0.853} & \underline{0.499}
& \textbf{12.36} & \underline{0.337} & \textbf{0.602} & \textbf{1.434} & \underline{0.765} \\
\bottomrule
\end{tabular}
}
\end{table*}

Our method outperforms existing baselines across most metrics, demonstrating the effectiveness of the proposed memory mechanism in preserving previously observed scene content during viewpoint revisiting. On RealEstate10K, it achieves the best PSNR, SSIM, LPIPS, and rotation error; on Tanks-and-Temples, it achieves the best PSNR, LPIPS, and rotation error. Translation error and Tanks-and-Temples SSIM remain better for some competing methods. Qualitative comparisons in Fig.~\ref{fig:comparison} further support these results. When returning to the starting viewpoint, \modelname better recovers scene details observed in the initial frame, while competing methods may exhibit missing content or structural inconsistencies.
\subsection{Generalization Across Scenes}
\label{sec:generalization}
The preceding comparisons evaluate the principal synthesis tasks on RealEstate10K and Tanks-and-Temples. We next examine whether the same memory framework supports scene preservation on DL3DV and on in-the-wild reference images.

\subsubsection{DL3DV}
\label{sec:dl3dv}
We evaluate \modelname and the competing methods on the 110 DL3DV clips described in the experimental setup, using the same viewpoint revisiting protocol and visual and geometric metrics as above. As shown in Table~\ref{tab:dl3dv_results}, \modelname achieves the best SSIM, LPIPS, and translation error among the compared methods. GEN3C has slightly higher PSNR, while VMem has lower rotation error. This comparison supports competitive generalization without implying uniform improvement across all metrics.

\begin{table}[t]
\centering
\caption{Quantitative comparison under the Viewpoint Revisiting
setting on the DL3DV dataset. Best results are highlighted in bold,
and second-best results are underlined.}
\label{tab:dl3dv_results}
\resizebox{\columnwidth}{!}{
\begin{tabular}{lccccc}
\toprule
Method
& PSNR $\uparrow$
& SSIM $\uparrow$
& LPIPS $\downarrow$
& $R_{\mathrm{dist}} \downarrow$
& $T_{\mathrm{dist}} \downarrow$ \\
\midrule
MotionCtrl~\cite{wang2024motionctrl}
& 9.52 & 0.274 & 0.735 & 2.090 & 1.082 \\

ViewCrafter~\cite{yu2024viewcrafter}
& 12.01 & 0.353 & 0.621 & 2.577 & 0.760 \\

GEN3C~\cite{ren2025gen3c}
& \textbf{12.41} & 0.340 & 0.587 & 2.147 & \underline{0.585} \\

SEVA~\cite{zhou2025stable}
& 12.14 & 0.326 & 0.609 & \underline{1.905} & 0.748 \\

VMem~\cite{li2025vmem}
& 12.28 & \underline{0.359} & \underline{0.550} & \textbf{1.792} & 0.632 \\

\midrule
\modelname~(Ours)
& \underline{12.35}
& \textbf{0.376}
& \textbf{0.522}
& 2.086
& \textbf{0.517} \\
\bottomrule
\end{tabular}
}
\end{table}

\begin{figure*}[t]
    \centering
    \includegraphics[width=\textwidth]{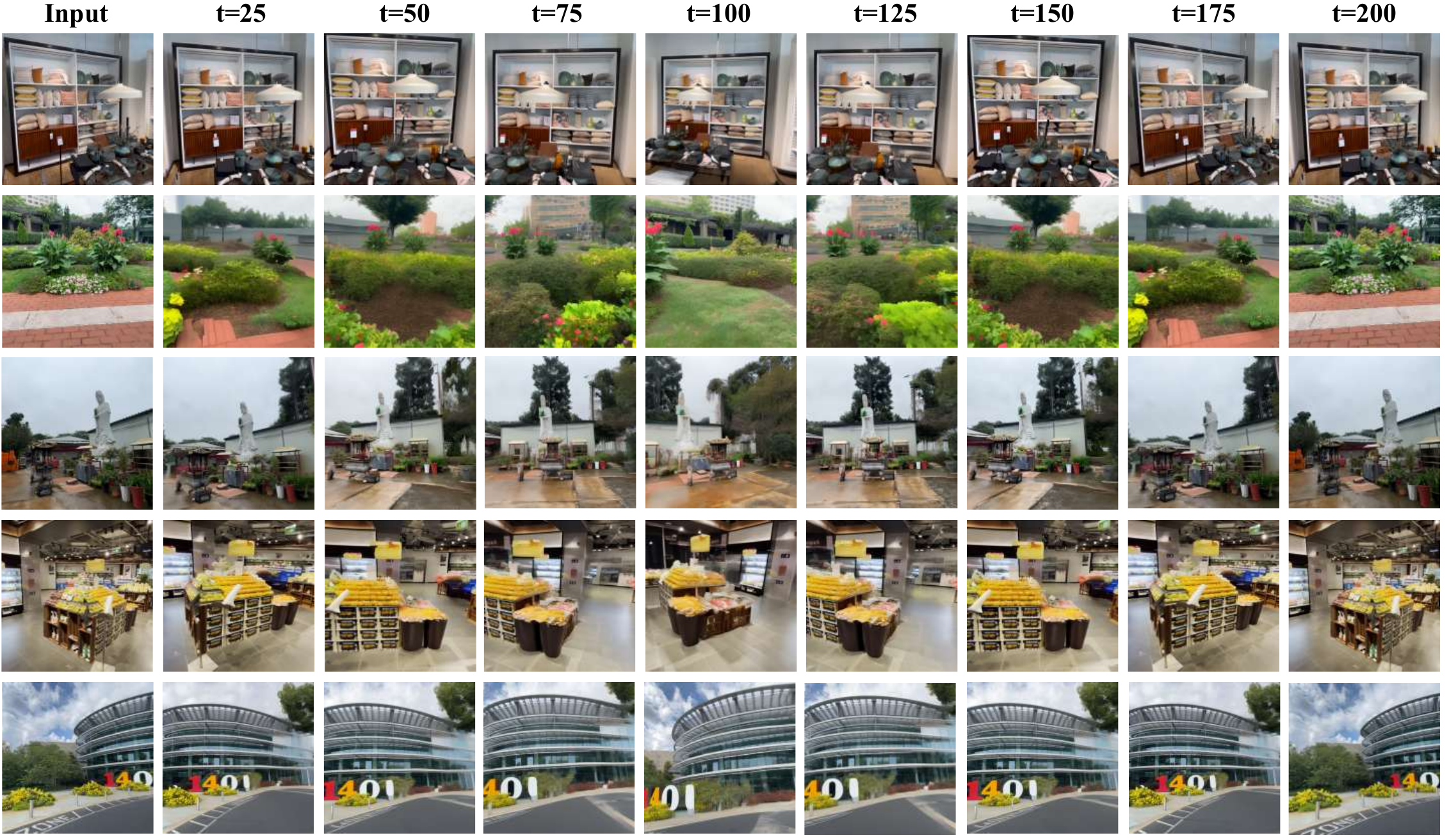}
    \caption{\textbf{Qualitative results on DL3DV under the Viewpoint Revisiting setting.}
    Representative examples show that \modelname preserves scene structure and previously observed content when the camera returns to earlier viewpoints across diverse indoor and outdoor scenes.}
    \label{fig:dl3dv_qualitative}
\end{figure*}

Fig.~\ref{fig:dl3dv_qualitative} shows representative indoor and outdoor examples in which \modelname preserves scene structure and previously observed content when the camera revisits earlier viewpoints. Together with the quantitative comparison, these results extend the evidence for scene consistency across diverse scenes and camera trajectories.

\subsubsection{In-the-Wild Synthesis}
\label{sec:wild}
\begin{figure*}[t]
    \centering
    \includegraphics[width=1.0\textwidth]{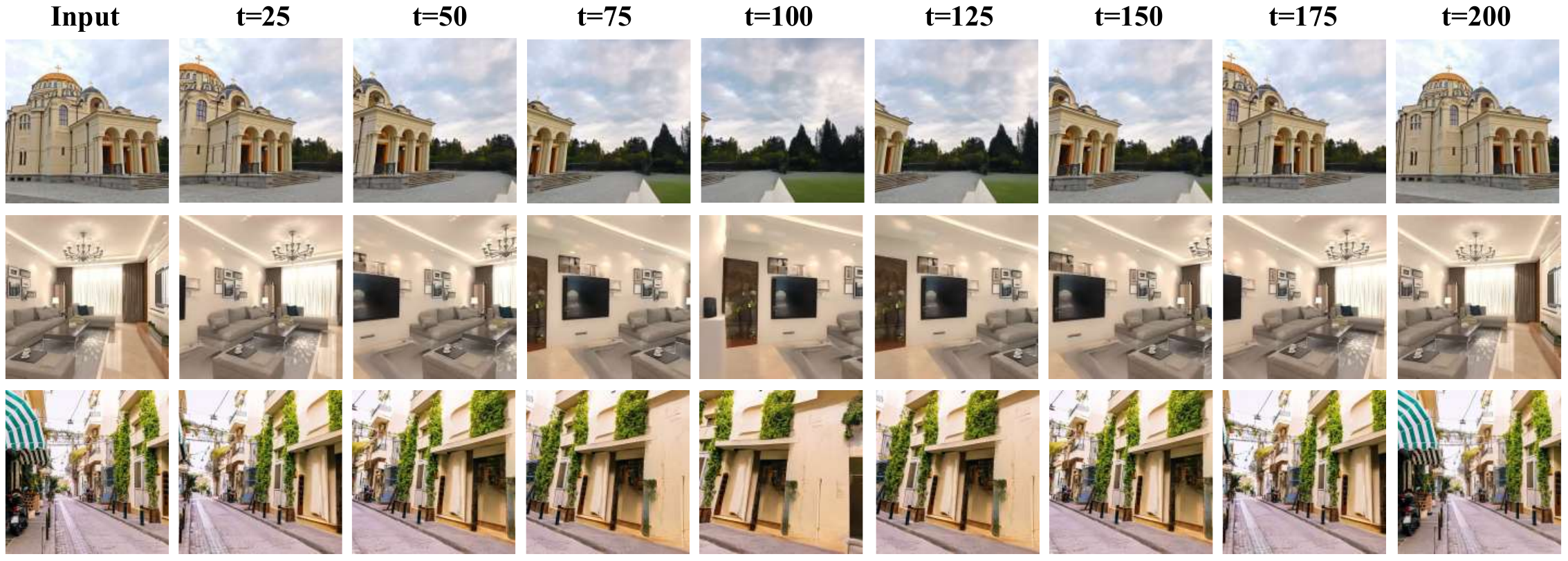}
    \caption{\textbf{Qualitative synthesis results on in-the-wild images from internet sources.}
    \modelname preserves scene structure across the illustrated camera trajectories without domain-specific fine-tuning.}
    \label{fig:wild}
\end{figure*}

To evaluate generalization beyond the benchmark scenes, we conduct experiments on in-the-wild images collected from internet sources, without domain-specific fine-tuning. Fig.~\ref{fig:wild} presents representative generated sequences with preserved scene structure across camera movement. These qualitative examples complement the benchmark measurements; no quantitative ground-truth comparison is reported for this setting.

\subsection{Ablation Studies}
\label{sec:ablations}
To validate our design choices, we conduct all ablation experiments under the viewpoint revisiting setting on the RealEstate10K dataset.

\subsubsection{Different Backbones}
To examine whether the benefit of \modelname extends beyond the default SEVA generator, we integrate it with SEVA~\cite{zhou2025stable}, ViewCrafter~\cite{yu2024viewcrafter}, and MVGenMaster~\cite{cao2025mvgenmaster}. For each backbone, the retrieved historical images and their camera parameters are supplied as conditioning observations through its existing interface. The memory module selects the context without changing the generator architecture, allowing the same retrieval strategy to be evaluated across the three backbones.

\begin{table}[t]
\centering
\caption{Effect of \modelname on different camera-conditioned backbones.}
\label{tab:backbone}
\resizebox{\linewidth}{!}{
\begin{tabular}{lcccccc}
\toprule
Backbone & Method & PSNR $\uparrow$ & SSIM $\uparrow$ & LPIPS $\downarrow$ & $R_{dist} \downarrow$ & $T_{dist} \downarrow$  \\
\midrule
{SEVA}
& Baseline & 13.20 & 0.501 & 0.459 & 0.967 & 0.516 \\
& + \modelname & \textbf{14.22} & \textbf{0.549} & \textbf{0.387} & \textbf{0.853} & \textbf{0.499} \\
\midrule
{ViewCrafter}
& Baseline & 13.38 & 0.473 & 0.434 &0.912 & 0.654  \\
& + \modelname & \textbf{13.90} & \textbf{0.490} & \textbf{0.414} & \textbf{0.727} & \textbf{0.536} \\
\midrule
{MVGenMaster}
& Baseline & 11.29 & 0.342 & 0.604 & 1.327 & 1.108  \\
& + \modelname & \textbf{13.25} & \textbf{0.472} & \textbf{0.479} & \textbf{1.056} & \textbf{0.942} \\
\bottomrule
\end{tabular}}
\end{table}

Table~\ref{tab:backbone} shows improvements in all five metrics for each backbone. PSNR increases from 13.20 to 14.22 for SEVA, from 13.38 to 13.90 for ViewCrafter, and from 11.29 to 13.25 for MVGenMaster. LPIPS and both pose errors also decrease in every case, indicating that the retrieved context benefits perceptual fidelity and geometric consistency together. The gains on ViewCrafter and MVGenMaster show that the usefulness of semantic-aware historical observations is not restricted to SEVA. These comparisons support the compatibility of the memory module with the tested generators.

\subsubsection{Memory Retrieval}
We investigate the effect of different retrieval strategies by comparing \modelname with three variants: 
(i) \textit{Sliding Window}, which selects recent temporal frames as context; 
(ii) \textit{FOV}, which follows the geometry-aware retrieval strategy introduced in Context-as-Memory~\cite{yu2025context}, selecting frames based on the field-of-view overlap between the target viewpoint and historical observations; 
and (iii) \textit{Ours (w/o semantic)}, which retains our 3D memory retrieval mechanism but removes semantic importance scoring. 
All variants are evaluated under the viewpoint revisiting setting. 
As shown in Table~\ref{tab:ablation}, our full model improves PSNR, SSIM, LPIPS, and rotation error relative to the variants, demonstrating the benefit of combining semantic importance with episodic memory retrieval for selecting informative historical observations. FOV has slightly lower translation error, indicating that the improvements are not uniform across every geometric metric.
\begin{table}[t]
\centering
\caption{Ablation study of memory retrieval strategy and semantic components on RealEstate10K.}
\label{tab:ablation}
\resizebox{\columnwidth}{!}{
\begin{tabular}{l ccccc}
\toprule
Variant & PSNR $\uparrow$ & SSIM $\uparrow$ & LPIPS $\downarrow$ & $R_{dist} \downarrow$ & $T_{dist} \downarrow$ \\
\midrule
(i) Sliding Window          & 13.05 & 0.496 & 0.472 & 1.120 & 0.637 \\
(ii) FOV                   & 13.70 & 0.516 & 0.430 & 0.904 & \textbf{0.488} \\
(iii) Ours (w/o semantic)  & 14.01 & 0.531 & 0.413 & 0.879 & 0.502 \\
\midrule
\textbf{(iv) Ours (w/ semantic)}  & \textbf{14.22} & \textbf{0.549} & \textbf{0.387} & \textbf{0.853} & 0.499 \\
\bottomrule
\end{tabular}
}
\end{table}

\subsubsection{Importance Scoring Components}
We analyze semantic distinctiveness, object-level information density, and local structural cues. As shown in Table~\ref{tab:importance}, object-level density improves both visual quality and geometric consistency, showing that feature diversity within object regions provides complementary information. Local structure further improves visual metrics by emphasizing fine-grained edges and textures. However, the variant without local structure yields slightly lower rotation and translation errors, suggesting that these cues benefit appearance preservation but may be less stable across viewpoint changes.

\begin{table}[t]
\centering
\small
\setlength{\tabcolsep}{3.5pt}
\caption{Ablation study of different importance scoring components on the viewpoint revisiting task.}
\label{tab:importance}
\resizebox{\linewidth}{!}{
\begin{tabular}{lccccc}
\toprule
Components 
& PSNR $\uparrow$ 
& SSIM $\uparrow$ 
& LPIPS $\downarrow$ 
& $R_{dist}\downarrow$ 
& $T_{dist}\downarrow$ \\
\midrule

Distinctiveness
& 13.71
& 0.526
& 0.421
& 0.921
& 0.561 \\

Distinctiveness + Object
& 14.15
& 0.542
& 0.392
& \textbf{0.824}
& \textbf{0.478} \\

Distinctiveness + Object + Local
& \textbf{14.22}
& \textbf{0.549}
& \textbf{0.387}
& 0.853
& 0.499 \\

\bottomrule
\end{tabular}
}
\end{table}

\subsubsection{Number of Complementary Frames}
\label{sec:ablation_n}
We evaluate the number of complementary frames under the same viewpoint revisiting setting on RealEstate10K. The total context budget is fixed at $K=4$, so increasing $N$ reduces the number of primary frames to $K-N$. As shown in Table~\ref{tab:ablation_n}, $N=1$ achieves the best performance across all reported metrics among the tested values $N\in\{0,1,2\}$. Compared with $N=0$, this supports the benefit of one complementary observation for informative regions missed by primary retrieval.

\begin{table}[t]
\centering
\caption{Ablation study on the number of complementary frames $N$ under the Viewpoint Revisiting setting on RealEstate10K. The total context window size is fixed at $K=4$.}
\label{tab:ablation_n}
\resizebox{\columnwidth}{!}{
\begin{tabular}{lccccc}
\toprule
$N$ & PSNR $\uparrow$ & SSIM $\uparrow$ & LPIPS $\downarrow$
& $R_{\mathrm{dist}} \downarrow$ & $T_{\mathrm{dist}} \downarrow$ \\
\midrule
$N=0$ & 13.97 & 0.534 & 0.391 & 0.885 & 0.508 \\
$N=1$ & \textbf{14.22} & \textbf{0.549} & \textbf{0.387}
& \textbf{0.853} & \textbf{0.499} \\
$N=2$ & 13.74 & 0.518 & 0.412 & 0.942 & 0.524 \\
\bottomrule
\end{tabular}
}
\end{table}

Increasing $N$ to $2$ degrades the reported metrics. Under the fixed budget, additional complementary frames replace primary observations selected for strong target-view support. The result is consistent with a tradeoff between complementary coverage and geometric relevance, rather than a benefit from increasing complementary context alone. We therefore use $N=1$ in the default configuration.
\subsection{Memory and Computational Cost}
\label{sec:efficiency}
We evaluate the computational overhead of \modelname under viewpoint revisiting, as summarized in Table~\ref{tab:efficiency}. All measurements are conducted on a single NVIDIA A100 GPU under the same experimental setting. The table reports peak GPU memory and generation time for the compared methods.

\begin{table}[t]
\centering
\caption{Memory and time cost under the viewpoint revisiting
setting. All measurements are conducted on a single NVIDIA A100 GPU.}
\small
\setlength{\tabcolsep}{10pt}
\begin{tabular}{lcc}
\toprule
\textbf{Method} & \textbf{Peak GPU Memory} & \textbf{Time (s)} \\
\midrule
SEVA & 17.84 GB & 175 \\
VMem          & 18.20 GB & 612 \\
\textbf{Ours} & 18.15 GB & 347 \\
\bottomrule
\end{tabular}
\label{tab:efficiency}
\end{table}

\modelname uses 18.15 GB of peak GPU memory, compared with 17.84 GB for SEVA and 18.20 GB for VMem. Its generation time is 347 s, exceeding the 175 s of SEVA but remaining below the 612 s of VMem. These measurements show the runtime cost of semantic importance estimation and memory retrieval together with a limited increase in peak GPU memory relative to the underlying generator.

Directly retaining point-level observations from every generated frame would continually add geometric records. Our voxel representation instead merges repeated observations through local attribute updates, so its geometric storage depends on occupied voxels. The reported peak-memory measurements characterize the evaluated trajectories; they do not establish constant total memory or runtime for arbitrary synthesis lengths, since exploring new regions and maintaining frame associations can enlarge the stored scene history.
\subsection{Discussion}
\label{sec:discussion}
\noindent\textbf{Reconstruction Error and Error Propagation.}
Point-map errors may affect the spatial associations and visibility estimates used for keyframe retrieval. However, \modelname conditions the generator on stored RGB frames rather than images rendered from the reconstructed point cloud. Thus, geometric reconstruction errors do not directly affect the visual content of the conditioning images.

The impact of geometric errors is therefore primarily on retrieval quality, as inaccurate spatial information may lead to the selection of less relevant frames. Confidence weighting helps mitigate this issue by reducing the influence of uncertain voxels. Nevertheless, our method does not eliminate artifacts in previously generated frames or prevent error accumulation during autoregressive generation.

%% file: files/conclusion.tex
\section{Conclusion}
We present \modelname, a Semantic-Episodic Memory framework for scene-consistent novel view synthesis along user-defined camera trajectories. It organizes accumulated 3D observations into episodic memory, augments them with semantic importance, and retrieves historical views through a two-stage strategy combining target-view relevance and complementary coverage. Experiments on \textit{RealEstate10K}, \textit{Tanks-and-Temples}, and \textit{DL3DV} support improved visual quality and scene consistency on most viewpoint revisiting metrics. Ablations support the importance cues and the balance between primary and complementary retrieval, while resource measurements quantify the additional cost over the base generator.
